\documentclass{article}
\usepackage{ijcai19}

\usepackage{times}
\usepackage{soul}
\usepackage{url}
\usepackage[hidelinks]{hyperref}
\usepackage[utf8]{inputenc}
\usepackage[small]{caption}
\usepackage{graphicx}
\usepackage{algorithm}
\usepackage{algorithmic}
\usepackage{amsfonts}

\usepackage{booktabs}
\usepackage{xspace}
\usepackage{enumitem}
\usepackage{array}
\usepackage{multirow}
\usepackage{amsmath}
\usepackage{subcaption}
\usepackage{tablefootnote}
\usepackage{balance}
\usepackage[export]{adjustbox}

\usepackage{hyperref}

\usepackage{fancyhdr}
\fancypagestyle{firstpage}
{
    \fancyhead[L]{Published as a conference paper at IJCAI 2019}    
}
\graphicspath{ {Figures/} }

\title{Node Embedding over Temporal Graphs}

\author{
Uriel Singer$^1$
\and
Ido Guy$^2$\And
Kira Radinsky$^1$
\affiliations
$^1$Technion, Israel Institute of Technology\\
$^2$eBay Research\\
\emails
urielsinger@cs.technion.ac.il,
idoguy@acm.org,
kirar@cs.technion.ac.il
}

\begin{document}

\maketitle

\newcommand{\specialcell}[2][c]{%
  \begin{tabular}[#1]{@{}c@{}}#2\end{tabular}}

\newcommand{\ig}[1]{\textcolor{blue}{$\ll$\textsf{#1 --IG}$\gg$}}
\newcommand{\us}[1]{\textcolor{red}{$\ll$\textsf{#1 --US}$\gg$}}
\newcommand{\kr}[1]{\textcolor{green}{$\ll$\textsf{#1 --KR}$\gg$}}

\newcommand\footnoteref[1]{\protected@xdef\@thefnmark{\ref{#1}}\@footnotemark}

\newcommand{\ntv}{{node2vec}\xspace}
\newcommand{\Ntv}{{Node2vec}\xspace}
\newcommand{\tntv}{{tNodeEmbed}\xspace}
\newcommand{\tmf}{{TMF}\xspace}
\newcommand{\tmfntv}{{TMFntv}\xspace}
\newcommand{\lst}{{TFNM}\xspace}
\newcommand{\listrain}{{LISTtrain}\xspace}
\newcommand{\ctdne}{{CTDNE}\xspace}
\newcommand{\ctdnel}{{CTDNE}\xspace}
\newcommand{\ctdnex}{{HTNE}\xspace}
\newcommand{\dt}{{DynTri}\xspace}

\newcommand{\wtv}{{word2vec}\xspace}

\newcommand{\argmin}{\mathop{\mathrm{argmin}}} 

\newcolumntype{L}[1]{>{\raggedright\let\newline\\\arraybackslash\hspace{0pt}}m{#1}}
\newcolumntype{C}[1]{>{\centering\let\newline\\\arraybackslash\hspace{0pt}}m{#1}}
\newcolumntype{R}[1]{>{\raggedleft\let\newline\\\arraybackslash\hspace{0pt}}m{#1}}
\begin{abstract}
In this work, we present a method for node embedding in temporal graphs. We propose an algorithm that learns the evolution of a temporal graph's nodes and edges over time and incorporates this dynamics in a temporal node embedding framework for different graph prediction tasks. We present a joint loss function that creates a temporal embedding of a node by learning to combine its historical temporal embeddings, such that it optimizes per given task (e.g., link prediction). The algorithm is initialized using static node embeddings, which are then aligned over the representations of a node at different time points, and eventually adapted for the given task in a joint optimization. We evaluate the effectiveness of our approach over a variety of temporal graphs for the two fundamental tasks of temporal link prediction and multi-label node classification, comparing to competitive baselines and algorithmic alternatives. Our algorithm shows performance improvements across many of the datasets and baselines and is found particularly effective for graphs that are less cohesive, with a lower clustering coefficient.

\end{abstract}

\section{Introduction}
\label{sec:intro}
Understanding the development of large graphs over time bears significant importance to understanding community evolution and identifying deviant behavior.
For example, identifying nodes that
have an unusual structure change over time might indicate an anomaly or fraud. Another structure change can help identify new communities in a social network and thus can be used to improve social recommendations of friends or detect roles in a network.
 
Many important tasks in static network analysis focus on predictions over nodes and edges. Node classification assigns probabilities to a set of possible labels. For example, a person's role in a social network. Another important task is link prediction, which involves assigning a probability to an edge. For example, predicting whether two users in a social network are friends. In this paper, we focus on dynamic predictions of future interactions that involve structural changes. For example, predicting a person's future role or whether two users will become friends next year. 

Classic techniques for node and link prediction aim to define a set of features to represent the vertices or edges. 
These are subsequently used in a supervised learning setting to predict the class.
Commonly, to learn the features, linear and
non-linear dimensionality reduction techniques such as Principal
Component Analysis (PCA) are applied. They transform the graph's adjacency matrix to maximize the variance
in the data. 
More recent approaches, which have shown substantial performance improvement, aim at optimizing an objective that preserves
local neighborhoods of nodes \cite{Perozzi:2014:DOL}.
Lately, embedding-based approaches 
\cite{Wang:2016:SDN} showed state-of-the-art performance for graph prediction tasks.
For example, \ntv~\cite{grover16node} optimizes for embedding, where nodes that share the same network community and/or similar roles have similar representations in a latent space of a lower dimension.
It performs biased random walks to generate neighbors of nodes, similarly to previous work in natural language processing \cite{word2vec}. 
The use of pre-computed embeddings as features for supervised learning algorithms allows to
generalize across a wide variety of domains and prediction tasks.

In this work, we extend the prior embedding-based approaches, to include the network's temporal behavior.
Intuitively, if \ntv mimics word embeddings as a representation for node embedding, we present an algorithm that mimics sentence embedding \cite{PalangiDSGHCSW15} as an extended representation of a node. Each word in the sentence is a temporal representation of the node over time, capturing the dynamics of its role and connectivity. We propose \tntv, a semi-supervised algorithm that learns feature representations for temporal networks.
We optimize for two objective functions: (1) preserving static network neighborhoods of nodes in a d-dimensional feature
space and (2) preserving network dynamics.
We present an optimization function to jointly learn both (1) and (2).
We achieve (1) by leveraging static graph embedding. Specifically, in this work we perform experiments with \ntv embeddings, which have the advantage of being unsupervised and scaling well over large graphs.
As graph embeddings do not preserve coherence, i.e., each training of node embedding by (1) on the same graph can provide different node embeddings, we explore several approaches for aligning the graph representation to only capture true network dynamics rather than stochasticity stemming from the embedding training.
We then achieve (2) by creating a final embedding of a node by a jointly learning how to combine a node's historical temporal embeddings, such that it optimizes for a given task (e.g., link prediction).

In our experiments, we present results for two types of predictions: (1) temporal link prediction, i.e., given a pair of disconnected nodes at time $t$, predict the existence of an edge between them at time $t{+}n$ ; (2) multi-label node classification, i.e., given a node, predict its class.
We experiment with a variety of real-world networks across diverse domains, including social, biological, and scholar networks.
We compare our approach with state-of-the-art baselines and demonstrate its superiority by a statistically significant margin.
In addition, we observe that our algorithm reaches better prediction performance on graphs with a lower clustering coefficient. We hypothesize that when the network is more cohesive, the contribution of the network dynamics to the prediction is lower. Intuitively, as graph generation follows the preferential attachment model \cite{Barabasi99emergenceScaling}, new nodes will tend to attach to existing communities rendering them more cohesive. The usage of dynamics is especially important in graphs that have not yet developed large cohesive clusters.
We show analysis on several synthetic graphs mimicking different growth dynamics to support this hypothesis.  
We also show that our alignment approach significantly improves the performance.

The contribution of this work is threefold:
(1) We propose \tntv, an algorithm for feature learning in temporal networks that optimizes for preserving both network structure and dynamics. We present results for both node classification and edge prediction tasks. 
(2) We study when network dynamics brings most value for network structure prediction. We identify that in order to learn it successfully, a minimum historical behavior of a node is needed. Additionally, we observe that predictions over graphs of higher clustering coefficients are not significantly improved by incorporating network dynamics into node embeddings.
(3) We empirically evaluate \tntv for both edge and node prediction on several real-world datasets and show superior performance.
\section{Related Work}
\label{sec:rw}
Various works have explored the temporal phenomena in graphs: \cite{Leskovec:2005:GOT} empirically studied multiple graph evolution process over time via the average node degree.
Other works studied social network evolution \cite{Leskovec:2008:KDD}, knowledge graph dynamics \cite{Trivedi:2017:KnowEvolveDT}, and information cascades on Facebook and Twitter \cite{Cheng:2014:WWW,Kupavskii:2012:CIKM}.

Temporal graph behavior has been studied in several directions:
some works \cite{Li:2016:ICLR,kipf2016semi} focused on temporal prediction problems where the input is a graph.
They studied numerous deep learning approaches for representing an entire graph and minimizing a loss function for a specific prediction task.
Other methods \cite{pei2016node,Xiaoyi:2014:SDM,yu17temporally} directly minimized the loss function for a downstream prediction task without learning a feature-representation.
Most of these methods do not scale
due to high training time requirements, as they build models per node~\cite{yu17temporally,Xiaoyi:2014:SDM} or models with parameters that depend on the amount of nodes \cite{Rakshit:arxiv:2018}. Other models do not scale well across multiple time-stamps \cite{Xiaoyi:2014:SDM} or scale well but at the cost of a relatively low performance~\cite{du2018dynamic}.
 
To overcome the scaling drawbacks, one of the common approaches today for node and edge embedding focuses on feature learning for graph prediction tasks.
Feature learning in static graphs is typically done by using the graph matrix representation and performing dimensionality reduction, such as spectral clustering or PCA \cite{Yan:2007:GEE}.
Yet, eigendecomposition of the graph matrix is computationally expensive, and therefore hard to scale for large networks. 

Most recent approaches \cite{Perozzi:2014:DOL,grover16node} for feature learning in graphs aim to find a representation for nodes by learning their embeddings via neural networks, with no need for manual feature extraction. These methods have shown state-of-the-art results for both node classification and edge prediction.
However, such static network models seek to learn properties
of individual snapshots of a graph. In this work, we extend the state-of-the-art feature-learning approach to include the temporal aspects of graphs. 
Some recent works have attempted to improve the static embeddings by considering historical embedding of the nodes and producing more stable static embeddings \cite{Goyal:2017:DynGEMDE}. In this work, we aim at leveraging the node and edge dynamics to yield better and more informative embeddings, which can later be used for temporal prediction tasks. Intuitively, rather than smoothing out the dynamics, we claim it brings high value for the embeddings and predictions.
Recent surveys provide good summaries and additional details~\cite{cui2018survey,goyal2018graph,hamilton2017representation}.

 Our approach is unique in providing an end-to-end architecture that can be jointly optimized to a given task. In addition, our evaluation shows the superiority of our approach for the relevant baselines over a variety of datasets for both the temporal link prediction and node classification tasks.
\section{Feature Learning Framework}
\label{sec:framework}
Let $G{=}(V,E)$ be a graph where each temporal edge $(u,v)_t \in E$ is directed from a node $u$ to a node $v$ at time $t$.
We define a temporal graph, $G_{t}=(V_{t},E_{t})$, as the graph of all edges occurring up to time $t$. 
We define the evolution of $G_{t}$ over time by the set of graph snapshots, $G_{t_1},...,G_{t_T}$, at $T$ different time steps, $t_1<...<t_T$. 
In this work, we propose an algorithm that learns the evolution of a temporal graph's nodes and edges over time in an end-to-end architecture for different prediction tasks.

Our goal is to find for each node $v \!\in\! V$ at time $T$ a feature vector $f_T(v)$ that minimizes the loss of any given prediction task. We consider two major prediction tasks: node classification and link prediction.
For the node classification task, we consider a categorical cross-entropy loss, i.e.:
\begin{center}	
$L_{task} = -\sum_{v\in V}{\log{Pr(class(v)|f_T(v))}}$
\end{center}	
where $class(v)$ is the class label of a node $v$.
For the link prediction task, we consider a binary classification loss, i.e.:
\begin{center}	
$L_{task} = -\sum_{v_1,v_2\in V}{\log{Pr((v_1,v_2)\in E_{t}|g(f_T(v_1),f_T(v_2)))}}$
\end{center}
where $g$ can be any function. In this work, we consider the concatenation function.
It is important to note that, without loss of generality, other tasks with corresponding loss functions can be supported in our framework.

We wish to leverage the set of graph snapshots, $G_{1},\ldots,G_{T}$ to learn a function $F_T$, s.t.: $f_T(v) = F_T(v, G_1,\ldots, G_T)$, which best optimizes for $L_{task}$.
To learn node dynamics, we formalize the embedding of a node $v$ at time $t+1$ in a recursive representation:
\begin{equation}\label{eq:combined_feature_vector}
f_{t+1}(v)=\sigma(A f_t(v) + B R_t Q_t v)
\end{equation}
where $f_0(v)=\vec 0$, $A,B,R_t$ and $Q_t$ are matrices that are learned during training, $v$ is a one-hot vector representing a node, and $\sigma$ is an activation function. We consider several such functions and further discuss them in Section~\ref{sec:combinetemporalembeding}.

We formulate the final temporal embedding learning problem as an optimization problem minimizing the following loss function: 
\begin{equation}\label{eq:main_loss}
L = \min_{A,B,Q_1,...Q_T,R_2,..R_T} L_{task}
\end{equation}
We optimize this equation using Adam over the model parameters defining the embedding $f_T$, i.e., $A,B,Q_1,\ldots,Q_T, R_2,..R_T$.

Intuitively, one can interpret $Q_t$ 
as a matrix of static node representation in a specific time $t$. 
Minimizing the loss for $Q_t$ can be thought of as optimizing for a node embedding of a static graph snapshot at a previous time point, such that the final $f_T$ optimizes $L_{task}$.
We consider several initialization mechanisms for $Q_t$.
Specifically, we learn the representation by sampling neighboring nodes from the corresponding graph $G_t$ at time $t$, for all nodes, while minimizing the following loss function:
\begin{equation}\label{eq:node2vec_loss}
-\sum_t\sum_{v_t\in V_t}{\log{Pr(N(v_t)|Q_t v_t)}}
\end{equation}
where $N(v)$ is a network of neighbors of a node $v$ generated through some network sampling strategy.
In Section~\ref{sec:nodeembedding}, we discuss in detail the initialization procedure. 

We also require that an alignment between consecutive time steps of the static embeddings is preserved by enforcing $R_t$ as a rotation matrix. 
$R_t$ therefore minimizes:
\begin{equation}\label{eq:alignment_loss}
\sum_t{\left \| R_{t+1}Q_{t+1} - Q_{t} \right \| + \lambda \|R_{t+1}^TR_{t+1}-I \|}
\end{equation}
where $R_1{=}I$ and $\lambda$ is a restriction. The first element ensures the embedded representations of a node between two consecutive time steps are similar, while the second element forces $R$ to be a rotation matrix. 
Section~\ref{sec:alignment} provides additional details about the alignment process.

This end-to-end procedure enables us to learn how to combine a node's historical temporal embeddings into a final embedding, such that it can optimize for a given task (e.g., link prediction), as defined by $L_{task}$. 
The rest of the section is structured as follows: we first discuss the initialization procedure for $Q_t$ (Section~\ref{sec:nodeembedding}). We then discuss in detail the optimization of $R_t$ (Section \ref{sec:alignment}). We finish the discussion of the framework by discussing the optimization using a deep learning architecture (Section \ref{sec:combinetemporalembeding}), which creates a final embedding of a node by learning how to combine its historical temporal embeddings, such that it optimizes per given task.

\subsection{Initialization using Static Node Embeddings}
\label{sec:nodeembedding}
In this section, we discuss the initialization procedure for $Q_t$.

Several prior works studied node embedding and reported good performance on \emph{static} graphs, i.e., graphs that either represent a specific point in time (snapshot) or do not change over time.
The architecture presented in this paper can use any of the known approaches for node embedding over static graphs. Specifically, we opted to work with \ntv, which achieved state-of-the-art performance on various benchmarks.

Due to the scalability of \ntv and its network preservation properties, we use it to initialize $Q_t$. Specifically, we compute the node embeddings for all $T$ graphs $G_{t_1},\ldots,G_{t_T}$. The outcome of this stage is a $T\times d$ vector per node, where $T$ is the number of time steps and $d$ is the embedding size.  
Those are used as initial values for $Q_t$, which will be further optimized for an end task.

\subsection{Temporal Node Embedding Alignment}
\label{sec:alignment}
In this section, we discuss in detail the optimization of the rotation matrix $R_t$.
Word2vec (and analogously \ntv) aims to minimize related word embedding distances, but does not guarantee embedding consistency across two distinct trainings.
Similarly, in the temporal graph domain, the embeddings of two graphs $G_{t_i}, G_{t_j}$ are performed independently, and therefore it is not guaranteed, even if the graphs are identical over the time points $t_i$ and $t_j$, that the node embeddings will remain the same. In other words, the coordinate axes of the embedded graph nodes are not guaranteed to align.

We aim to ``smooth out'' the variation between two different time steps $t_i$ and $t_j$ that originate from different embedding training sessions.
Assuming that most nodes have not changed much between $t_i$ and $t_j$ (e.g., $j{=}i{+}\epsilon$), we preform an orthogonal transformation between the node embeddings at time $t_i$ and the node embeddings at time $t_j$. Specifically, we use Orthogonal Procrustes~\cite{hurley1962procrustes}, which performs a least-squares approximation of two matrices.
Applied to our problem, let $Q_{t}\!\in\! \mathbb{R}^{d\times|V|}$ be the matrix of node embeddings at time step $t$.
We align the matrices corresponding to consecutive time steps iteratively, i.e., align $Q_{t_2}$ to $Q_{t_1}$, followed by aligning $Q_{t_3}$ to $Q_{t_2}$ and so forth.
The alignment requires finding the orthogonal matrix $R$ between $Q_{t}$ and $Q_{t+1}$. The final embedding at time $Q_{t}$ is now $Q'_{t} = RQ_{t}$.
The approximation is performed by optimizing the following regression problem:
\begin{center}	
$R_{t+1} = \argmin_{R \text{ s.t. } R^TR=I} \left \| RQ_{t+1} - Q_{t} \right \|$
\end{center}
where $R_{t+1} \in \mathbb{R}^{d\times d}$ is the best orthogonal transformation alignment between the two consecutive time steps.
Notice the regression problem is performed on nodes that appear both at time $t$ and time $t{+}1$. New nodes that only appeared at $t{+}1$ are transformed using $R_{t{+}1}$. 

\subsection{Node Embedding over Time}
\label{sec:combinetemporalembeding}
In eq.~\ref{eq:combined_feature_vector}, the final embedding is dependent on matrices $A, B$, and an activation function $\sigma$, which are jointly optimized as described in eq.~\ref{eq:main_loss}. In this section, we discuss the choice of $A, B$ and $\sigma$ and the final joint optimization.

Following previous steps, each node, $v$, is now associated with a matrix $X^{(v)} \in \mathbb{R}^{T\times d}$ of its historical $T$ embeddings over time, each  of size $d$.
The graph $G$ is associated with $|V|$ matrices, one for each node: $G_X = X^{(v_1)},\ldots,X^{(v_{|V|})}$.
Given $G_X$, we wish to perform graph prediction tasks -- node classification and link prediction.

The common approach for these tasks is to represent a node via a $d$-dimensional vector, which can then be used as input to any supervised machine learning-classifier.  
Similar to our problem but in the text domain, a sentence consists of a set of words, each with an associated embedding of size $d$. 
For the task of sentence classification, each sentence is embedded into a a vector of size $d$, which is then fed into a classifier.

Analogously, in this work, at the final steps of the optimization, we aim to create for each node a single $d$-dimensional representation, leveraging its $T$ historical embeddings, $X^{(v)}$.
In order to reduce sequence data into a $d$-dimensional vector, we use recurrent neural networks (RNNs) with long short term memory (LSTM).

\begin{figure}
\centering
\small
\begin{subfigure}{.2\textwidth}
	\hspace*{-0.45cm} 
  \includegraphics[width=1.30\linewidth]{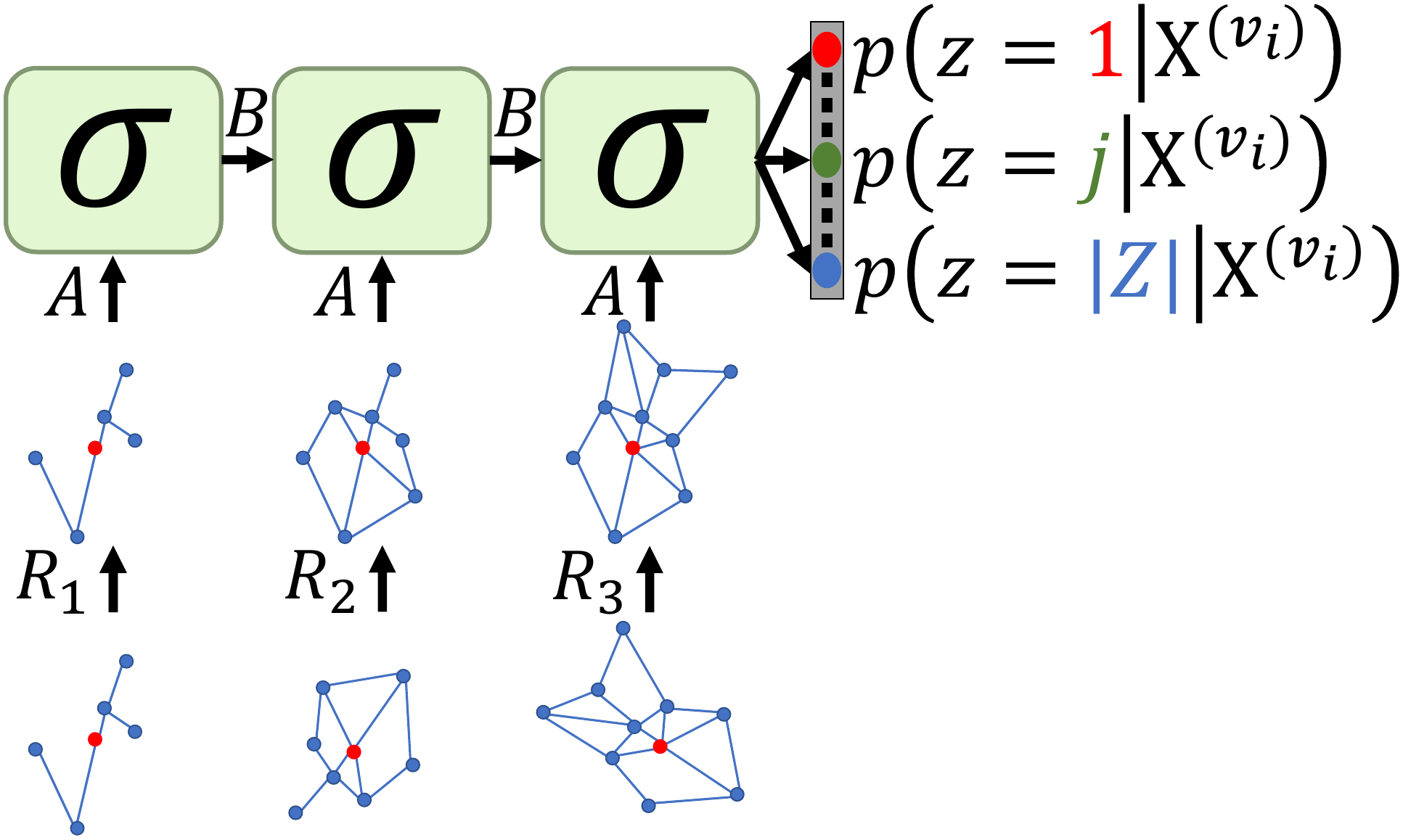}
  \caption{}
\label{Figure:node_architecture}
\end{subfigure}\qquad
\begin{subfigure}{.2\textwidth}
    \hspace*{-0.02cm} 
  \includegraphics[width=1.10\linewidth]{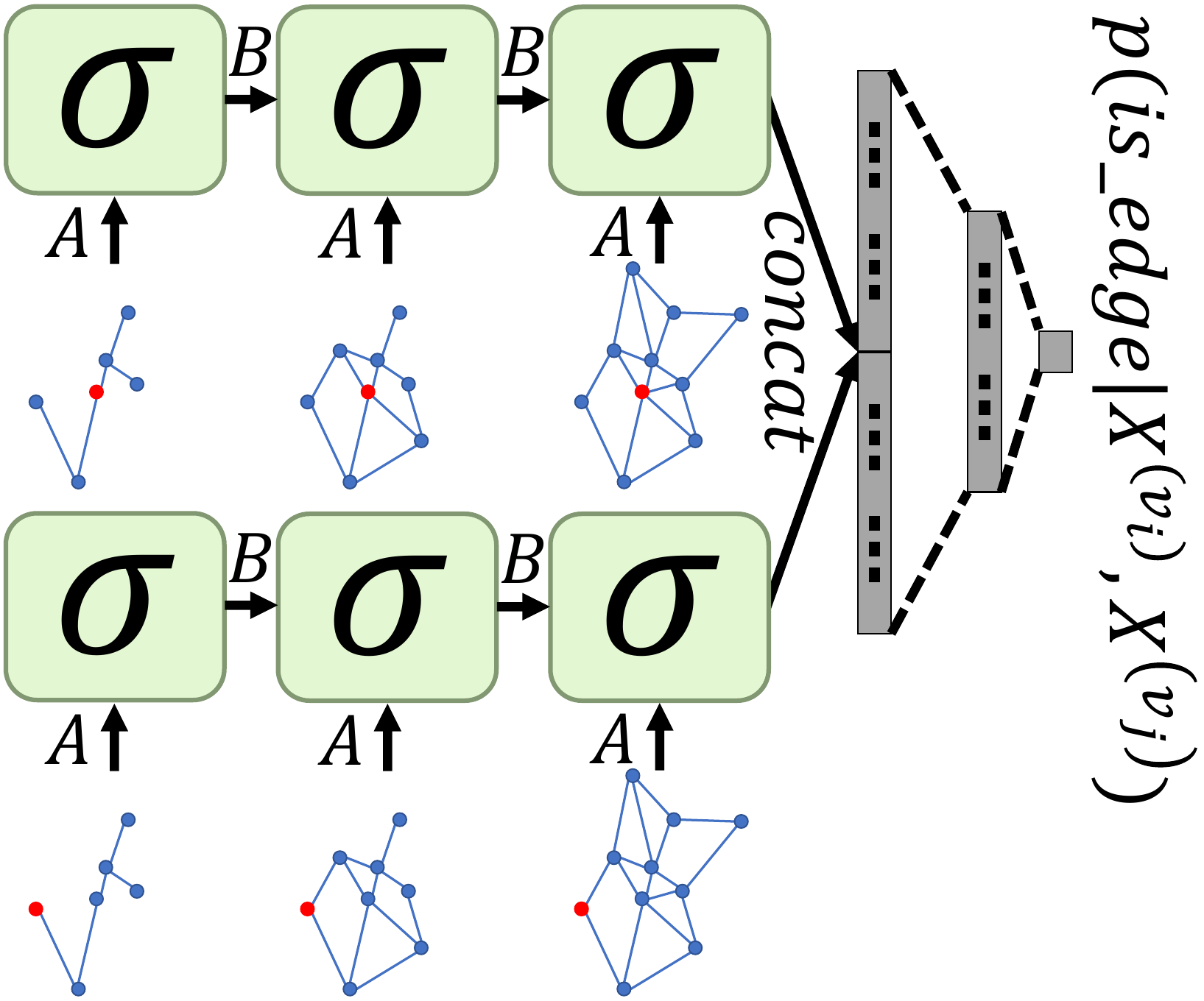}
\caption{}
\label{Figure:link_architecture}
\end{subfigure}
\caption{\normalsize{End-to-end architecture for node classification (\subref{Figure:node_architecture}) and link prediction (\subref{Figure:link_architecture}) (with no alignment).
}}
\label{Figure:architecture}
\end{figure}

We define \tntv as the architecture that is initialized using static node embedding and alignment, and is optimized for a given task. 
Figure~\ref{Figure:node_architecture} illustrates the process for the multi-label node classification task. Each node embedding at time $t$, after alignment, is fed as input to an LSTM memory cell of size $d$. The last memory cell $h_T \in \mathbb{R}^{d}$ of the LSTM  represents the final temporal embedding of the node, $v^t$, optimizing for the specific task. For the link prediction task (Figure~\ref{Figure:link_architecture}), each edge $e=(v_1,v_2)$ is represented by $(v^t_1, v^t_2)$ -- a vector of size $2d$, which is the concatenation of the two nodes it connects. This edge representation is then fed into a fully-connected (FC) layer of size $d$, followed by a softmax layer. Intuitively, The FC layer tries to predict the existence of the potential edge. 

\section{Experimental Setup}
\label{sec:experiments}

In this section, we describe our datasets, our two experimental tasks, and our baselines, which include a static embedding and previously-published temporal graph prediction algorithms.\footnote{\label{code_footnote}We publicly publish our code and data: \url{https://github.com/urielsinger/tNodeEmbed}}

\subsection{Datasets}
Table~\ref{table:datasets} lists the different datasets we used for our experiments and their characteristics.
Below, we describe each of them in more detail.

\paragraph{arXiv hep-ph.}
A research publication graph, where each node is an author and a temporal undirected edge represents a common publication with a timestamp of the publication date. Time steps reflect a monthly granularity between March 1992 and December 1999.

\paragraph{Facebook friendships.}
A graph of the Facebook social network where each node is a user and temporal undirected edges represent users who are friends, with timestamps reflecting the date they became friends. Time steps reflect a monthly granularity between September 2004 and January 2009.

\paragraph{Facebook wall posts.}
A graph of the Facebook social network where each node is a user and a temporal directed edge represents a post from one user on another user's wall at a given timestamp. Time steps reflect a monthly granularity between September 2006 and January 2009.

\paragraph{CollegeMsg.}
An online social network at the University of California, with users as nodes and a temporal directed edge representing a private message sent from one user to another at a given timestamp. Time steps reflect a daily granularity between April 15th, 2004 and October 26th, 2004.

\paragraph{PPI.}
The protein-protein interactions (PPI) graph includes protein as nodes, with edges connecting two proteins for which a biological interaction was observed. Numerous experiments have been conducted to discover such interactions. These are listed in HINTdb~\cite{patil2005hint}, with the list of all articles mentioning each interaction. We consider the interaction discovery date as the edge's timestamp. In a pre-processing step, we set it as the earliest publication date of its associated articles. We work in a yearly granularity between $1970$ and $2015$. We publicly release this new temporal graph.\textsuperscript{\ref{code_footnote}}

\paragraph{Slashdot.}
A graph underlying the Slashdot social news website, with users as nodes and edges representing replies from one user to another at a given timestamp. Time steps reflect a monthly granularity between January 2004 and September 2006.

\paragraph{Cora.}
A research publication graph, where each node represents a publication, labeled with one of $L{=}10$ topical categories: artificial intelligence, data structures algorithms and theory, databases, encryption and compression, hardware and architecture, human computer interaction, information retrieval, networking, operating systems, and programming. Temporal directed edges represent citations from one paper to another, with timestamps of the citing paper's publication date. Time steps reflect a yearly granularity between 1900 and 1999.

\paragraph{DBLP.}
A co-authorship graph, focused on the Computer Science domain. Each node represents an author and is labeled using conference keywords representing $L{=}15$ research fields: verification testing, computer graphics, computer vision, networking, data mining, operating systems, computer-human interaction, software engineering, machine learning, bioinformatics, computing theory, security, information retrieval, computational linguistics, and unknown. Temporal undirected edges represent co-authorship of a paper, with timestamps of the paper's publication date.  Time steps reflect a yearly granularity between 1990 and 1998.
\vskip 3pt

Notice that for the arXiv hep-ph, Facebook Wall Posts, CollegeMsg, Slashdot, and DBLP, multiple edges may occur from one node to another at different timestamps. Given a timestamp, if multiple edges exist, they are collapsed into a single edge, weighted according to the number of original edges, thus rendering a weighted graph, as marked in Table~\ref{table:datasets}.

\begin{table}[tb]
\centering
\scriptsize
 \setlength{\tabcolsep}{0.25em}
  \begin{tabular}
  {@{}lC{1.0cm}C{0.9cm}C{0.8cm}C{1.0cm}C{1.0cm}C{0.8cm}}
\toprule
Dataset & Weighted & Directed & Nodes & Edges & Diameter & Train time steps \\
\midrule
{arXiv hep-ph}\tablefootnote{\url{http://konect.uni-koblenz.de/networks/ca-cit-HepPh}} & + & - & 16,959 & 2,322,259 & 9 & 83 \\ 
{Facebook friendships}\tablefootnote{\url{http://konect.uni-koblenz.de/networks/facebook-wosn-links}} & - & - & 63,731 & 817,035 & 15 & 26 \\ 
{Facebook wall posts}\tablefootnote{\url{http://konect.uni-koblenz.de/networks/facebook-wosn-wall}} & + & + & 46,952 & 876,993 & 18 & 46\\
{CollegeMsg}\tablefootnote{\url{https://snap.stanford.edu/data/CollegeMsg.html}} & + & + & 1,899 & 59,835 & 8 & 69 \\
PPI\textsuperscript{\ref{code_footnote}}& - & - & 16,458 & 144,033 & 10 & 37 \\
{Slashdot}\tablefootnote{\url{http://konect.uni-koblenz.de/networks/slashdot-threads}} & + & + & 51,083 & 140,778 & 17 & 12 \\ 
{Cora}\tablefootnote{\url{https://people.cs.umass.edu/~mccallum/data.html}} & - & + & 12,588 & 47,675 & 20 & 39 \\ 
{DBLP}\tablefootnote{\url{http://dblp.uni-trier.de/xml}} & + & - & 416,204 & 1,436,225 & 23 & 9 \\
\bottomrule
\end{tabular}
\caption{\label{table:datasets} Dataset characteristics. 
}
\end{table}

\subsection{Experimental Tasks}
\label{sec:tasks}
We evaluated our temporal node embedding approach with regards to two fundamental tasks: temporal link prediction and multi-label node classification.
For link prediction, the first six datasets in Table~\ref{table:datasets} were used, while for node classification the remaining two datasets, Cora and DBLP, which include node labels, were used. 

\subsubsection{Temporal Link Prediction}
This task aims at forecasting whether an edge will be formed between two nodes in a future time point.

\paragraph{Data.}
We divided the data into train and test by selecting a pivot time, such that $80\%$ of the edges in the graph (or the closest possible portion) have a timestamp earlier or equal to the pivot. Following, all edges with an earlier (or equal) timestamp than the pivot time were considered as positive examples for the training set, while all edges with a timestamp later than the pivot time were considered as positive examples for the test set. For negative examples in the training set, we sampled an identical number of edges to the positive examples, uniformly at random out of all node pairs not connected at pivot time. For negative examples in the test set, we sampled uniformly at random an identical number of edges to the positive examples, from all node pairs not connected by an edge at all. We focused our task on predicting ties between existing nodes and therefore restricted edges in the test set to be formed only between existing nodes in the training set.\footnote{The selection of the pivot time took into account this restriction.}

\paragraph{Metric.}
As evaluation metric for all methods, we used the area under the ROC curve (AUC).

\subsubsection{Multi-Label Node Classification} 
This task aims at predicting the label of a node out of a given set of $L$ labels.

\paragraph{Data.}
For this task, we randomly split the entire dataset so that $80\%$ of the nodes are used for training.

\paragraph{Metrics.}
As our main evaluation metric, we used the F1 score. We examined both the micro F1 score, which is computed globally based on the true and false predictions, and the macro F1 score, computed per each class and averaged across all classes. For completeness, we also report the AUC.

\subsection{Baselines}
\label{sec:baselines}
In our task evaluations, we used the following baselines: 

\paragraph{\Ntv.}
We use \ntv as the state-of-the-art static baseline. It is also the static node embedding algorithm we opted to use for the initialization of \tntv, therefore the comparison between them is of special interest.\footnote{For this baseline, we used the implementation published by the authors: \url{https://github.com/aditya-grover/node2vec}}

\paragraph{Temporal Matrix Factorization (\tmf).}~\cite{dunlavy2011temporal}
This method, used specifically for link prediction, collapses the data across multiple times into a single representation over which matrix factorization is applied. In addition, we also experiment with a setting in which \ntv is used on top of the collapsed matrix, as a more modern method for embedding, and mark this variant as \tmfntv.

\paragraph{Temporally Factorized Network Modeling (\lst).}~\cite{yu17temporally}
This method applies factorization before collapsing, by generalizing the regular notion of matrix factorization for matrices with a third `time' dimension.\footnote{For this baseline, we used the authors implementation kindly shared with us.}

\paragraph{Continuous-Time Dynamic Network Embeddings (\ctdne).}~\cite{nguyen2018continuous}
This method is based on random walks with a stipulation that the timestamp of the next edge in the walk must be larger than the timestamp of the current edge.

\paragraph{Hawkes process-based Temporal Network Embedding (\ctdnex).}~\cite{zuo2018embedding}
This method works similarly to \ctdne, but with neighborhoods generated by modeling Hawkes processes~\shortcite{hawkes1971spectra}, where each edge is weighted exponentially by the time difference. For \ctdnex and \ctdne, we used the finest time granularity included in each dataset.

\paragraph{DynamicTriad (\dt).}~\cite{zhou2018dynamic}
This method uses the modeling of the triadic closure process to learn representation vectors for nodes at different time steps.\footnote{For this baseline, we used the implementation published by the authors: \url{https://github.com/luckiezhou/DynamicTriad}}
\vskip 3pt

For all baselines, link prediction and node classification are implemented by using the node embeddings of the last time step (which holds all the graphs data). Classification is performed as described in Section~\ref{sec:combinetemporalembeding}.
\section{Experimental Results}
\label{sec:results}

The principal part of our evaluation includes a detailed comparison of \tntv with all baselines described in the previous section for the link prediction and node classification tasks. We then further examine the performance for the link prediction task over four types of random graphs, with different degree distributions.
We conclude with an analysis of the effect of our alignment step on performance of both tasks.
Throughout this section, boldfaced results indicate a statistically significant difference.

\subsection{Temporal Link Prediction}
\begin{table*}
\centering
  \small
 \setlength{\tabcolsep}{0.12mm}
 \renewcommand{\arraystretch}{1.15}
  \begin{tabular}
 {@{}lC{1.7cm}C{1.6cm}C{1.6cm}C{1.6cm}C{1.6cm}C{1.6cm}C{1.6cm}C{1.6cm}C{1.5cm}@{}}
\toprule
Dataset & \tntv & \ntv & \tmf & \tmfntv & \lst & \ctdnel & \ctdnex & \dt & CC\\
\midrule
arXiv hep-ph & $0.951$ & $0.948$ & $0.908$ & $0.950$ & $0.738$ & $0.905$ & $0.851$ & $0.783$ & $0.291$\\ 
Facebook friendships & $0.939$ & $0.938$ & $0.886$ & $0.925$ & $0.814$ & $0.757$ & $0.724$ & $0.535$ & $0.148$\\ 
Facebook wall posts & $\textbf{0.917}$ & $0.902$ & $0.718$ & $0.900$ & $0.720$ & $0.827$ & $0.784$ & $0.643$ & $0.078$\\
CollegeMsg & $0.841$ & $0.823$ & $0.809$ & $0.819$ & $0.654$ & $0.841$ & $0.838$ & $0.630$ & $0.036$\\
PPI & $\textbf{0.828}$ & $0.799$ & $0.753$ & $0.798$ & $0.712$ & $0.800$ & $0.782$ & $0.761$ & $0.017$\\
Slashdot & $\textbf{0.913}$ & $0.777$ & $0.896$ & $0.793$ & $0.661$ & $0.894$ & $0.886$ & $0.765$ & $0.010$\\ 
\bottomrule
\end{tabular}
\caption{\label{table:tempoal_link_auc} AUC performance for link prediction of \tntv vs. baselines. Clustering coefficient is presented at the rightmost column.
}
\end{table*}

Table~\ref{table:tempoal_link_auc} presents the performance results of \tntv compared to the different baselines for the link prediction task. It can be seen that \tntv outperforms all other baselines across three datasets (Facebook wall posts, PPI, and Slashdot) and reaches comparable results in the other three. 
It also poses the most consistent performance, achieving the highest result for all six datasets. The performance results as well as the gap from the baselines vary substantially across the datasets. For arXiv hep-ph and Facebook friendships, \ntv and \tmfntv (our own composed baseline combining temporal matrix factorization with \ntv embedding) achieve comparable results to \tntv. For CollegeMsg, both \ctdnel and \ctdnex achieve comparable results to \tntv. Interestingly, it can be noticed that the static \ntv baseline outperforms \tmf and \lst across all datasets, with the exception of Slashdot for \tmf. This indicates the strength of modern node embedding methods and suggests that the temporal data across all time steps is not guaranteed to yield a performance gain on top of such embedding. The poor results of \lst may stem from the fact we examine substantially larger graphs. It also implies that the assumption of a first-order polynomial tie across time points may be too restrictive in such cases. The particularly low performance demonstrated by \dt may stem from the fact that the loss in each time step uses only data from the current and previous time steps. As a result, the broader dynamics of the graph are not captured as effectively as in other methods.
Indeed, the authors demonstrated the effectiveness of this method for tasks that use the node representations in time $t$ to make predictions for time $t$ or $t{+}1$, but not in further steps, where the entire dynamics of the graph become important to capture.  

To better understand on which types of graphs \tntv is most effective, we analyzed several graph properties and 
discovered a particularly consistent correlation with the global \textit{clustering coefficient} (CC)~\shortcite{wasserman94social}.
Intuitively, the CC reflects how well nodes in a graph tend to cluster together, or how \textit{cohesive} the graph is.
As Table~\ref{table:tempoal_link_auc} indicates, \tntv tends to perform better compared to the baselines for graphs with a lower CC.
We conjecture the reason lies in the cohesiveness of the graph over time. As most graphs follow the preferential attachment model, nodes tend to attach to existing communities, rendering them more cohesive.
The ``denser'' the graph, the easier it is to predict the appearance of a link, as edges tend to attach to higher degree nodes.

\subsection{Multi-Label Node Classification}
Table~\ref{table:node_classification} presents the results for the node classification task. In this case, \tntv outperforms all other baselines across all metrics over both datasets, except for the case of micro $F_1$ for DBLP, which is equal to that of \lst. However, for the latter, the macro $F_1$ is especially low, implying a collapse into one label. A similar phenomenon of a low macro $F_1$ can be observed for \ctdnel, \ctdnex, and \dt. This suggests that the time scale of years in both the Cora and DBLP datasets is not suitable for continuous methods. In addition, it reinforces our conjecture that \dt may work well only for predictions for the current or next time step.  

It can also be seen that as opposed to the link prediction task, the performance gaps between \tntv and the baselines are rather similar for Cora and DBLP, despite the CC difference between the two graphs. Indeed, by contrast to link prediction, node classification is not directly related to the cohesiveness of the graph. 

Overall, our evaluation indicates that \tntv achieves clear performance gains over the baselines for both the link prediction and node classification tasks.

\begin{table}[tb]
\centering
\scriptsize
\setlength{\tabcolsep}{0.15em}
\renewcommand{\arraystretch}{1.15}
\begin{tabular}
	{@{}lcccccc@{}}
\toprule
\multirow{2}{*}{Algorithm} & \multicolumn{3}{c}{Cora (CC=0.275)} & \multicolumn{3}{c}{DBLP (CC=0.002)}\\
\cmidrule(r){2-4}\cmidrule(l){5-7}
& Micro $F_1$ & Macro $F_1$ & AUC & Micro $F_1$ & Macro $F_1$ & AUC\\
\midrule
\tntv & $\textbf{0.668}$ & $\textbf{0.513}$ & $\textbf{0.925}$ & $0.822$ & $\textbf{0.504}$ & $\textbf{0.977}$\\
\ntv & $0.547$ & $0.284$ & $0.862$ & $0.752$ & $0.235$ & $0.943$\\
\tmf & $0.552$ & $0.361$ & $0.875$ & $0.740$ & $0.203$ & $0.937$\\
\tmfntv & $0.511$ & $0.246$ & $0.856$ & $0.749$ & $0.219$ & $0.939$\\
\lst & $0.386$ & $0.078$ & $0.760$ & $0.822$ & $0.060$ & $0.937$\\
\ctdnel & $0.374$ & $0.054$ & $0.753$ & $0.717$ & $0.083$ & $0.916$\\
\ctdnex & $0.391$ & $0.056$ & $0.747$ & $0.714$ & $0.069$ & $0.911$\\
\dt & $0.386$ & $0.055$ & $0.746$ & $0.711$ & $0.055$ & $0.897$\\
\bottomrule
\end{tabular}
\caption{\label{table:node_classification} Performance results of \tntv vs. baselines for the node classification task over the Cora and DBLP datasets.
}
\end{table}

\subsection{Degree Distribution} 
\label{generating}
In order to further explore the performance of \tntv on different types of graphs, and the aforementioned effect of the CC, we superficially generated four random graphs with different degree distributions other than power law: linear, logarithmic, sinusoidal, and exponential. All graphs were created for $n{=}1000$ nodes and $m{=}100{,}000$ edges. We increased the number of edges linearly along $T{=}50$ time steps, so that the total number of edges after all $T$ time steps would be $m$. At each time step $t$, we added $m_t$ new random edges to the graph, between its $n$ nodes, so that the degree distribution would be as close as possible to the predefined degree distribution. We ran \tntv, \ntv, and \ctdnel for the link prediction task, using $t{=}45$ as the pivot. We opted for \ctdnel as a representative temporal baseline, since it achieved the best performance over $3$ of the $6$ datasets for the temporal link prediction task (as can be seen in table ~\ref{table:tempoal_link_auc}). Notice that since the edge selection is not completely random, but has to follow a predefined degree distribution, node embedding is still expected to be meaningful. 

Table~\ref{table:random} shows the performance results, which are consistent with those observed for real-world graphs with a power-law degree distribution. For all four distributions, \tntv outperforms \ntv and is comparable with \ctdnel. As previously observed, performance increases as the CC of the graph grows, while the gap of \tntv over \ntv is larger when the CC is lower. The similar results for \tntv and \ctdnel algorithms imply that the temporal dynamics for these graphs is relatively easy to capture.

\begin{table}
\centering
\small
  \begin{tabular}
  {@{}lcccc@{}}
\toprule
Distribution & \tntv & \ntv & \ctdnel & CC \\
\midrule
Linear & $0.67$ & $0.53$ & $0.65$ & $0.22$ \\
Logarithmic & $0.74$ & $0.56$ & $0.73$ & $0.26$ \\ 
Sinusoidal & $0.79$ & $0.64$ & $0.78$ & $0.31$ \\
Exponential & $0.85$ & $0.79$ & $0.83$ & $0.36$ \\
\bottomrule
\end{tabular}
\caption{\label{table:random} 
AUC performance for temporal link prediction over randomly generated graphs with different degree distribution targets.
}
\end{table}

\subsection{Alignment}
\label{subsec:alignment}
As explained in Section~\ref{sec:alignment}, \tntv aligns node embeddings over different time points. This alignment aims to learn consistent node behavior over time and reduce noise that arises from training different embeddings.
To examine the effect of the alignment step, we experimented with a variant of \tntv that skips this step. Table~\ref{table:link_alignment} presents the results of this variant alongside the results of the ``full-fledged'' \tntv for the temporal link prediction task. It can be seen that the removal of the alignment step leads to a decrease in performance in three out of six datasets, while in the rest the performance is comparable.
Overall, these results indicate that the alignment step can play a key role in performance enhancement.
We observe that the lower the CC of the graph, the higher the contribution of the alignment step.
We conjecture that the embeddings variability constructed between each embedding trainings is higher for graphs that are less cohesive. This generates more noise in the data, which leads to a lower performance of \tntv. The alignment step helps reduce this noise by aligning the embeddings across time points. The results for the node classification task follow the same trends and are presented in Table~\ref{table:node_classification_alignment}.

\begin{table}
\centering
  \small
  \begin{tabular}
  {@{}lccc@{}}
\toprule
Dataset & Alignment & No Alignment & CC \\
\midrule
arXiv hep-ph & $0.951$ & $0.950$ & $0.29$ \\
Facebook friendships & $0.939$ & $0.935$ & $0.15$ \\ 
Facebook wall posts & $0.917$ & $0.916$ & $0.08$ \\
CollegeMsg & $\textbf{0.841}$ & $0.825$ & $0.04$ \\
PPI & $\textbf{0.828}$ & $0.822$ & $0.02$ \\
Slashdot & $\textbf{0.913}$ & $0.860$ & $0.01$ \\
\bottomrule
\end{tabular}
\caption{\label{table:link_alignment} AUC performance for link prediction using \tntv with and without alignment.
}
\end{table}

\begin{table}
\centering
\small
  \setlength{\tabcolsep}{0.37em}
  \begin{tabular}
	{@{}lccccccc@{}}
\toprule
\multirow{2}{*}{Dataset} & \multicolumn{2}{c}{Micro $F_1$} & \multicolumn{2}{c}{Macro $F_1$} & \multicolumn{2}{c}{AUC} & \multirow{2}{*}{CC}\\
\cmidrule(r){2-3}\cmidrule(lr){4-5}\cmidrule(l){6-7}
& with & without & with & without & with & without & \\
\midrule
Cora & $\textbf{0.668}$ & $0.644$ & $\textbf{0.513}$ & $0.475$ & $0.925$ & $0.919$ & $0.275$\\
DBLP & $\textbf{0.822}$ & $0.785$ & $\textbf{0.504}$ & $0.390$ & $\textbf{0.977}$ & $0.959$ & $0.002$\\
\bottomrule
\end{tabular}
\caption{\label{table:node_classification_alignment} Performance for node classification using \tntv with and without alignment.
}
\end{table}
\section{Conclusions}
\label{sec:conclusions}
In this paper, we explore temporal graph embeddings. 
Unlike previous work in predictions over temporal graphs, which focused on optimizing one specific task, we present a framework that allows to jointly optimize the node representations and the task-specific objectives. Our method outperforms a variety of baselines across many datasets and does not underperform for any of them. We evaluate our method over a particularly diverse set of large-scale temporal graphs, weighted and unweighted, directed and undirected, with different time spans and granularities, for two fundamental graph prediction tasks -- node classification and edge prediction. 

Our method generalizes graph embedding to capture graph dynamics, somewhat similarly to the extension of word embedding to sequential sentence embedding in natural language processing. Our framework can leverage any static graph embedding technique and learn a temporal node embedding altering it.

As future work, we would like to extend our framework to learn the embeddings by using the best time resolution, without taking snapshots at discrete time points.

\bibliographystyle{named}
\bibliography{tnode}

\end{document}